\UseRawInputEncoding

\documentclass[letterpaper, 10 pt, conference]{ieeeconf}  

\IEEEoverridecommandlockouts                              

\usepackage{cite}
\usepackage{amsmath,amssymb,amsfonts}
\usepackage{algorithmic}
\usepackage{graphicx}
\usepackage{textcomp}
\usepackage{xcolor}
\usepackage{booktabs}
\usepackage{algorithm}  
\usepackage{algorithmic}

\title{\LARGE \bf
Speed and Density Planning for a Speed-Constrained Robot Swarm Through a Virtual Tube
}

\author{Wenqi Song, Yan Gao and  Quan Quan$^*$ 
\thanks{Wenqi Song, Yan Gao, Quan Quan (Corresponding Author) are with School of Automation Science and Electrical Engineering, Beihang University, Beijing, 100191, P.R. China {\tt\small \{buaa\_swq, buaa\_gaoyan, qq\_buaa\}}{\tt\small @buaa.edu.cn}}%
}

\begin{document}

\maketitle
\thispagestyle{empty}
\pagestyle{empty}

\begin{abstract}

The planning and control of a robot swarm in a complex environment have attracted increasing attention. To this end, the idea of virtual tubes has been taken up in our previous work. Specifically, a virtual tube with varying widths has been planned to avoid collisions with obstacles in a complex environment. Based on the planned virtual tube for a large number of speed-constrained robots, the average forward speed and density along the virtual tube are further planned in this paper to ensure safety and improve efficiency. Compared with the existing methods, the proposed method is based on global information and can be applied to traversing narrow spaces for speed-constrained robot swarms. Numerical simulations and experiments are conducted to show that the safety and efficiency of the passing-through process are improved. A video about simulations and experiments is available on https://youtu.be/lJHdMQMqSpc.

\end{abstract}


\section{Introduction}
Swarm planning and control in a complex environment have attracted more and more attention. The main goal is to find an optimal route for each robot without collisions with other robots and obstacles from the starting point to the goal subject to the kinematic conditions. How to make the robot swarm pass through complex environments more safely and faster is an important issue that researchers are constantly exploring \cite{b40}.

Many methods have been proposed for the passing-through problem of the robot swarm in a complex environment. For example, formation control\cite{b3,b41} strives to maintain a pre-determined rigid shape while traversing cluttered environments. 
In addition, multi-robot trajectory planning algorithms \cite{b7} are widely used to plan a geometric path for the robot in a swarm that does not conflict with obstacles and other robots \cite{b42,b43}. 
Furthermore, control-based methods have also been proposed and widely used \cite{b18}. Classical methods include artificial potential field method \cite{b44}, vector field method, control barrier function method \cite{b8}, etc.
However, these methods may fail when a large number of robots are to pass through some narrow spaces. In this case, the robustness and scalability of formation control are limited, the calculation amount of multi-robot trajectory planning increases dramatically, and control-based methods easily leads to congestion. Moreover, multi-robot trajectory planning methods depend on direct communication heavily.

\begin{figure}[!t]
	\centering
	\setlength{\abovecaptionskip}{-0.3cm}
	\includegraphics[width=1\linewidth]{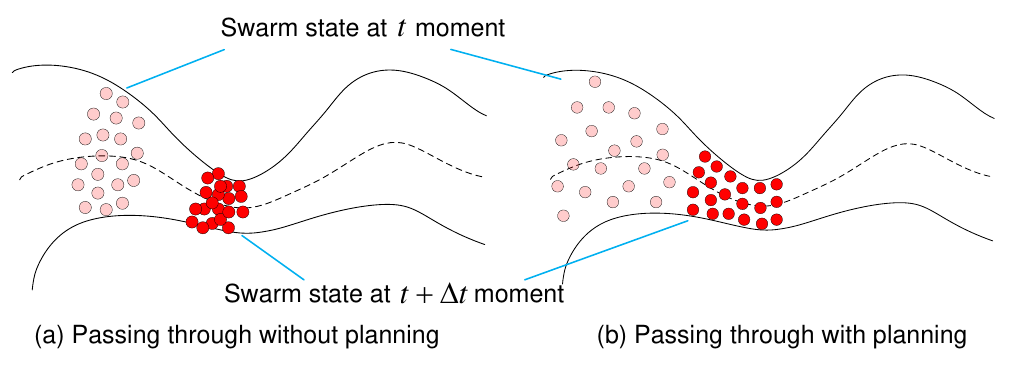}
	\caption{Comparison of swarm distribution in a virtual tube without planning and with planning. Light red indicates the swarm distribution before dark red.}
	\label{fig:f220909_12}
	\vspace{-0.6cm}
\end{figure}

For such a purpose, control within a virtual tube is proposed \cite{b17,b21}, where all robots sharing one planned virtual tube are under distributed control. The virtual tube can be seen as a safety corridor, which means there are no obstacles inside the virtual tube. This idea is natural and intuitive because it is similar to cars sharing one road under distributed control by human drivers. In the cluttered environment, there are always some narrow spaces, which provides a narrowing virtual tube.
However, for the speed-constrained robots, such as fixed-wing unmanned aerial vehicle (UAV), the swarm cannot stop to avoid colliding with each other when entering the narrowing virtual tube based on our previous control method \cite{b17}. This brings serious safety risks. In addition, congestion is possible to occur during the large-scale swarm passing-through process, which slows down the speed of the swarm. Therefore, effective advance planning is required to optimize the control. 

The problem of generating a collision-free passage for a speed-constrained swarm in complex environments is studied widely \cite{b31}. For example, the decentralized flocking with obstacle avoidance policy is learned for multiple fixed-wing UAVs based on a multiagent deep reinforcement learning approach \cite{b32}. Consider the planning of swarms, speed planning is mostly designed to optimize the speed of robots as well as save energy \cite{b33}. The main idea of speed planning is to present a path and speed planner under the physical constraints of the robot \cite{b34}. In addition, density planning is an effective method to ensure the safety of the swarm. For instance, a density planner is designed to generate a trajectory with the minimum collision probability under dynamic obstacles based on the initial distribution \cite{b35}. Furthermore, the optimal control problem is solved to make the swarm quickly converge to the desired density distribution \cite{b36}.

In this paper, we conduct the speed and density planning as well as traversing control for a speed-constrained robot swarm based on the established narrowing virtual tube to pass through some narrow spaces. Here, density planning is used to plan appropriate densities at different positions of the tube, so as to avoid collisions on account of the inability to stop. Meanwhile, the average forward speed along the virtual tube is planned to ensure efficiency. Then, distributed control is performed on each individual to track the planning results.

The contributions of this paper are as follows.
\begin{itemize}
	\item A new approach is proposed to solve the passing-through problem for a speed-constrained robot swarm within a narrowing virtual tube, which is full of challenges. This approach consists of a planned virtual tube, speed and density planning, and distributed control, where the latter two are considered. 
\end{itemize}
\begin{itemize}
	\item Speed and density planning is applied to control within a virtual tube \emph{for the first time}, which brings a tradeoff between the improvement of efficiency and safety. Moreover, speed and density planning for the whole swarm rather than an individual is very suited for a larger number of robots.
\end{itemize}

\section{PROBLEM FORMULATION}

\subsection{Robot Modeling}

\subsubsection{Robot Kinematic Model}

A robot is set up as a two-dimensional mass point model with speed constraints. The swarm is composed of $N$ homogeneous robots. In the Cartesian coordinate system, the motion model of the $i$th robot is
\begin{equation}
	\setlength\abovedisplayskip{3pt}
	\setlength\belowdisplayskip{3pt}
	{{{\mathbf{\dot{p}}}}_{i}}=	{{\mathbf{v}}_{\text{c},i}},
	\label{v00}		
\end{equation}
where $i=1,2,\cdots ,N$, ${{\mathbf{p}}_{i}}\in {{\mathbb{R}}^{2}}$ represents the position of the $i$th robot,  $	{{\mathbf{v}}_{\text{c},i}} \in {{\mathbb{R}}^{2}}$ represents the velocity command of the $i$th robot, $N$ represents the number of the robots in the swarm.

According to the mobility limitations of robots, the robots are restricted by the maximum speed ${{v}_\text{max }}$, the minimum speed  ${{v}_\text{min }}$,  the maximum tangential acceleration ${{a}_\text{v}}$,  and the maximum normal acceleration ${{a}_\text{n}}$ as follows: 
\begin{equation}
	\setlength\abovedisplayskip{3pt}
	0<{{v}_\text{min }}\le \left\| {{\mathbf{v}}_{\text{c},i}} \right\|\le {{v}_\text{max }},
	\label{v0}		
\end{equation}
\begin{equation}
	\setlength\abovedisplayskip{2pt}
	\setlength\belowdisplayskip{2pt}
	\frac{\text{d}\left\| {{\mathbf{v}}_{\text{c},i}} \right\|}{\text{d}t}\le {{a}_\text{v}},
	\label{v1}
\end{equation}
\begin{equation}
	\setlength\abovedisplayskip{1pt}
	\setlength\belowdisplayskip{1pt}
	\frac{{{\left\| {{\mathbf{v}}_{\text{c},i}} \right\|}^{2}}}{{{r}_{\text{t}}}}\le {{a}_\text{n}},
	\label{v2}
\end{equation}
where ${{r}_{\text{t}}}>0$ represents the radius of the robot's trajectory curvature.

\subsubsection{Physical Area, Safety Area and Obstacle Avoidance Area of Robots}

\begin{figure}[!t]
	\centering
	\setlength{\abovecaptionskip}{-0.1cm}
	\includegraphics[width=0.8\linewidth]{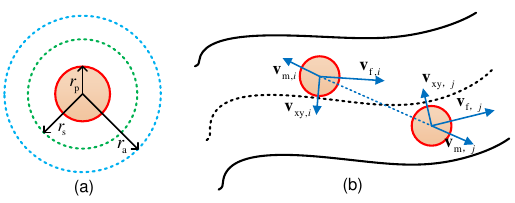}
	\caption{(a) The physical area, safety area, obstacle avoidance area of a robot. (b) The velocity command of the $i$th robot and the $j$th robot.}
	\label{fig:f_add}
	\vspace{-0.3cm}
\end{figure}

Concentric circles of different sizes are used to represent the physical area, safe area, and obstacle avoidance area of robots. As shown in Fig. \ref{fig:f_add} (a), ${{r}_{\text{p}}},{{r}_{\text{s}}},{{r}_{\text{a}}}$ denote the radius of the physical area, the safety area,  and the obstacle avoidance area respectively. Besides, there exists ${{r}_{\text{p}}}\!\le {{r}_{\text{s}}}\!\le {{r}_{\text{a}}}$ \cite{b17}. Particularly, ${{r}_{\text{a}}}$ is a controlled variable corresponding to the planned swarm density in this paper.
To be specific, for the $i$th robot, a controller is set up to track the planned swarm density by changing the magnitude of ${{r}_{\text{a},i}}$, that is ${{\dot{r}}_{\text{a},i}}\!={{r}_{\text{ac},i}}$, where ${{r}_{\text{ac},i}}$ is a controller for the radius of the obstacle avoidance area, which will be specified later.

\subsection{Virtual Tube Modeling}



\begin{figure}[!t]
	\centering
	\setlength{\abovecaptionskip}{-0.1cm}
	\includegraphics[width=0.65\linewidth]{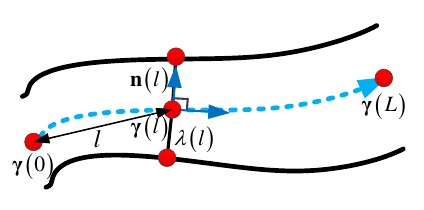}
	\caption{Schematic diagram of a virtual tube.}
	\label{fig:ff99}
	\vspace{-0.5cm}
\end{figure}

A virtual tube is a \emph{regular} curved tube designed on a two-dimensional plane \cite{b12}. As shown in Fig. \ref{fig:ff99}, the virtual tube in a two-dimensional plane is expressed as:
\begin{equation*}
	\setlength\abovedisplayskip{3pt}
	\setlength\belowdisplayskip{3pt}
	\mathcal{J } \left( l,\theta ,\rho  \right)=\boldsymbol{\gamma }\left( l \right)+\rho \lambda \left( l \right)\mathbf{n}\left( l \right)\cos \theta,		
\end{equation*}
in which $\theta =\left\{ 0,\pi  \right\}$, $l\in \left[ 0,L \right]$, $\rho \in \left[ {0},{1} \right]$. The curve $\boldsymbol{\gamma }\left( l \right)$ is the \emph{generator curve (center curve)} of the virtual tube. The vector $\mathbf{n}\left( l \right)$  represents the normal vector of the generator curve, $l$ represents the arc length of the generator curve from the starting point $\boldsymbol{\gamma }\left( 0 \right)$, and $\boldsymbol{\gamma }\left( l \right)$ is the position with the arc length $l$ along the generator curve from $\boldsymbol{\gamma }\left( 0 \right)$. Furthermore, $L>0$ represents the whole length of the generator curve, that is, the arc length from the starting point denoted by $\boldsymbol{\gamma }\left( 0 \right)$ to the ending point denoted by $\boldsymbol{\gamma }\left( L \right)$.  Moreover, $\lambda \left( l \right)$ is continuous, which represents the widths of the virtual tube. Additionally, ${{r}_{\text{t} }}(l)$ represents the curvature radius of the tube center curve. The detailed virtual tube generation theories and methods are introduced in our previous work \cite{b12}. In this paper, the position of the $i$th robot within the virtual tube is defined as ${{\mathbf{p}}_{i}}=\mathcal{J }\left( {{l}_{i}},{{\theta }_{i}},{{\rho }_{i}} \right)$. Particularly, each ${{\mathbf{p}}_{i}}$ corresponds to a unique $ {{l}_{i}}$, where ${l}_{i}\in \left[ 0,L \right]$.

\subsection{Robot Controller}

In this paper, the movement of the robot is controlled by velocity command. As shown in Fig. \ref{fig:f_add} (b), the velocity command ${{\mathbf{v}}_{\text{c},i}}$ of the $i$th robot consists of three components: ${{\mathbf{v}}_{\text{f},i}}$ to guide the robot to move forward along the virtual tube, ${{\mathbf{v}}_{\text{m},i}}$ to prevent the conflict among robots, and ${{\mathbf{v}}_{\text{xy},i}}$ to restrict the robot in the virtual tube. Particularly, ${\left\|{\mathbf{v}}_{\text{c},i}\right\|}\ne 0$. Our previous work \cite{b17} describes the controller design in detail. Here it is omitted for space limitation. For the $i$th robot, the velocity command is
\begin{equation}
	\setlength\abovedisplayskip{3pt}
	\setlength\belowdisplayskip{3pt}
	{{\mathbf{v}}_{\text{c},i}}=\text{sat}\left( {{\mathbf{v}}_{\text{f},i}}+{{\mathbf{v}}_{\text{m},i}\left( {{r}_{\text{a},i}} \right)}+{{\mathbf{v}}_{\text{xy},i}\left( {{r}_{\text{a},i}} \right)},{{v}_\text{min }},{{v}_\text{max }} \right).
	\label{s1}
\end{equation}
Here,
\begin{equation}
	{{\mathbf{v}}_{\text{f},i}}={{v}_{\text{f},i}}{{\mathbf{t}}_{\text{c}}}\left( {{\mathbf{p}}_{i}} \right),
	\label{a1}
\end{equation}
\begin{equation*}
	\text{sat}\left( \mathbf{v},{{v}_\text{min }},{{v}_\text{max }} \right)\triangleq \left\{ \begin{matrix}
		{{v}_\text{min }}\frac{\mathbf{v}}{\left\| \mathbf{v} \right\|} & \left\| \mathbf{v} \right\|< {{v}_\text{min }}  \\
		\mathbf{v} & {{v}_{\min }}\le \left\| \mathbf{v} \right\|\le {{v}_\text{max }}  \\
		{{v}_\text{max }}\frac{\mathbf{v}}{\left\| \mathbf{v} \right\|} & \left\| \mathbf{v} \right\|> {{v}_\text{max }}  \\
	\end{matrix} \right..
\end{equation*}
where ${{{v}}_{\text{f},i}}$ represents the modulus of ${{\mathbf{v}}_{\text{f},i}}$, and ${{\mathbf{t}}_{\text{c}}}\left( {{\mathbf{p}}_{i}} \right)$ represents the tangent vector of the projection point of the $i$th robot on the virtual tube center curve. Particularly, ${\left\|{\mathbf{v}}_{\text{m},i}\left( {{r}_{\text{a},i}} \right)\right\|}$ and ${\left\|{\mathbf{v}}_{\text{xy},i}\left( {{r}_{\text{a},i}} \right)\right\|}$ will be increased at the moment when ${{r}_{\text{a},i}}$ is increased. In other words, the swarm will be expanded like gas expansion after heating.

\textbf{Remark 1.}
When a robot is modeled as a single integrator such as (\ref{v00}), exemplified by some holonomic kinematics robots such as multicopters, helicopters, and specific variants of omni-directional wheeled robots, the designed velocity command ${{\mathbf{v}}_{\text{c},i}}$ can be straightforwardly employed to control the robot.
When dealing with a more complicated model, such as a second-order integrator model, additional control laws become imperative. In our previous work \cite{b17}, we introduced a \emph{filtered position model} that transforms a second-order model into a first-order model just like (\ref{v00}). As for certain nonholonomic kinematics robots such as ground mobile robots and fixed-wing UAVs, we can further generate appropriate forward speed command or angular speed command tailored to the model. This process ensures that the robot velocity can track the designed velocity command ${{\mathbf{v}}_{\text{c},i}}$, that is,  ${{\lim }_{t\to \infty }}\left\| {{\mathbf{v}}_{i}}\left( t \right)-{{\mathbf{v}}_{\text{c},i}}\right\|=0$ \cite{b26}. Another approach involves the utilization of a near-identity diffeomorphism to establish a connection between the desired single integrator model and the more precise robot model \cite{b45}.

\subsection{Density and Average Forward Speed}

\begin{figure}[!t]
	\centering
	\setlength{\abovecaptionskip}{0cm}
	\includegraphics[width=0.8\linewidth]{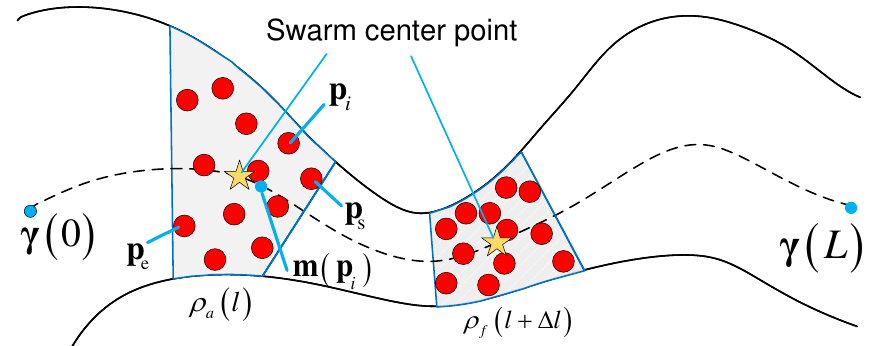}
	\caption{Distributions of the swarm in the virtual tube at the position with arc length $l$ and $\left(l + \Delta l \right)$.}
	\label{fig:pf}
	\vspace{-0.3cm}
\end{figure}

In this paper, the swarm density ${{\rho }_{\text{a}}}$ is defined as the number of robots in a unit area,
\begin{equation}
	\setlength\abovedisplayskip{1pt}
	\setlength\belowdisplayskip{3pt}
	{{\rho }_{\text{a}}} ={N}/{S},
	\label{p0}
\end{equation}
where $S$ is the area occupied by the swarm within the virtual tube. The area $S$ is the gray area in Fig. \ref{fig:pf}. Assume that the swarm passes through the virtual tube from the starting point $\boldsymbol{\gamma }\left( 0 \right)$ to the ending point $\boldsymbol{\gamma }\left( L\right)$. Let ${{\mathbf{p}}_{\text{e}}}$ denote the position of the robot farthest away $\boldsymbol{\gamma }\left( L\right)$ in the swarm (the last robot). And let ${{\mathbf{p}}_{\text{s}}}$ denote the position of the robot nearest to $\boldsymbol{\gamma }\left( L\right)$ in the swarm (the front robot). Besides, let $\mathbf{m}\left( {{\mathbf{p}}_{i}} \right)$ denote the projection of the $i$th robot on the tube center curve. Then, the position of the front robot in the swarm is
\begin{equation*}
	\setlength\abovedisplayskip{3pt}
	\setlength\belowdisplayskip{3pt}
	{{\mathbf{p}}_\text{s}}=\underset{{{\mathbf{p}}_{i}}}{\mathop{\arg \min }}\,s\left( \mathbf{m}\left( {{\mathbf{p}}_{i}} \right),\boldsymbol{\gamma }\left( L \right) \right),
\end{equation*}
where $s\left( \mathbf{m}\left( {{\mathbf{p}}_{i}} \right),{\boldsymbol{\gamma }\left( L \right)} \right)$ denotes the arc length of the tube center curve between $\mathbf{m}\left( {{\mathbf{p}}_{i}} \right)$ and $\boldsymbol{\gamma }\left( L \right)$. Similarly, the position of the last robot is
\begin{equation*}
	\setlength\abovedisplayskip{3pt}
	\setlength\belowdisplayskip{3pt}
	{{\mathbf{p}}_\text{e}}=\underset{{{\mathbf{p}}_{i}}}{\mathop{\arg \max }}\,s\left( \mathbf{m}\left( {{\mathbf{p}}_{i}} \right),\boldsymbol{\gamma }\left( L \right) \right).
\end{equation*}
Thus the area occupied by the swarm is
\begin{equation}
	\setlength\abovedisplayskip{3pt}
	\setlength\belowdisplayskip{3pt}
	S=-\int_{\mathbf{m}\left( {{\mathbf{p}}_\text{e}} \right)}^{\mathbf{m}\left( {{\mathbf{p}}_\text{s}} \right)}{2{\lambda}\left( \mathbf{p} \right)}\text{d}s\left( \mathbf{p},{\boldsymbol{\gamma }\left( L \right)} \right).
	\label{ps}
\end{equation}


\textit{\textbf{Assumption 1.}}
The swarm is considered as a single point called the swarm center point with average speed ${{v}_\text{a}}$ and swarm density ${{\rho}_\text{a}}$.

Based on \textit{Assumption 1}, the swarm center point is the yellow pentagram in Fig. \ref{fig:pf}. The forward speed of this point during the passing-through process is planned. The average forward speed of the swarm is
${{v}_{\text{a}}}=\frac{1}{N}\sum\nolimits_{i=1}^{N}{{{v}_{\text{f},i}}}$.

\subsection{Problem Formulation}

In this paper, a speed-constrained swarm moving in a complex environment is simplified as moving within a virtual tube with varying width $\lambda \left( l \right)$. The goal is to ensure the passing-through \emph{safety} and \emph{efficiency}. This virtual tube is supposed to be pre-designed. Let ${{v}_\text{a}^{*}}(l)$ and $ {{\rho }_\text{a}^{*}}(l)$ denote the planned average forward speed and the planned density respectively.

\textit{\textbf{Assumption 2.}}
The area occupied by a robot is a circumscribed square of its circular obstacle avoidance area. Moreover, the area occupied by the swarm is minimum when the circumscribed square of robots' circular safety area are closely adjacent.

\textit{\textbf{Assumption 3.}}
The swarm moves forward along the center curve without relative position change of any pair of robots, which means that the projection on the tube center curve of each robot moves the same distance along the tube center curve.

\begin{figure}[!t]
	\centering
	\setlength{\abovecaptionskip}{-0.2cm}
	\includegraphics[width=0.35\linewidth]{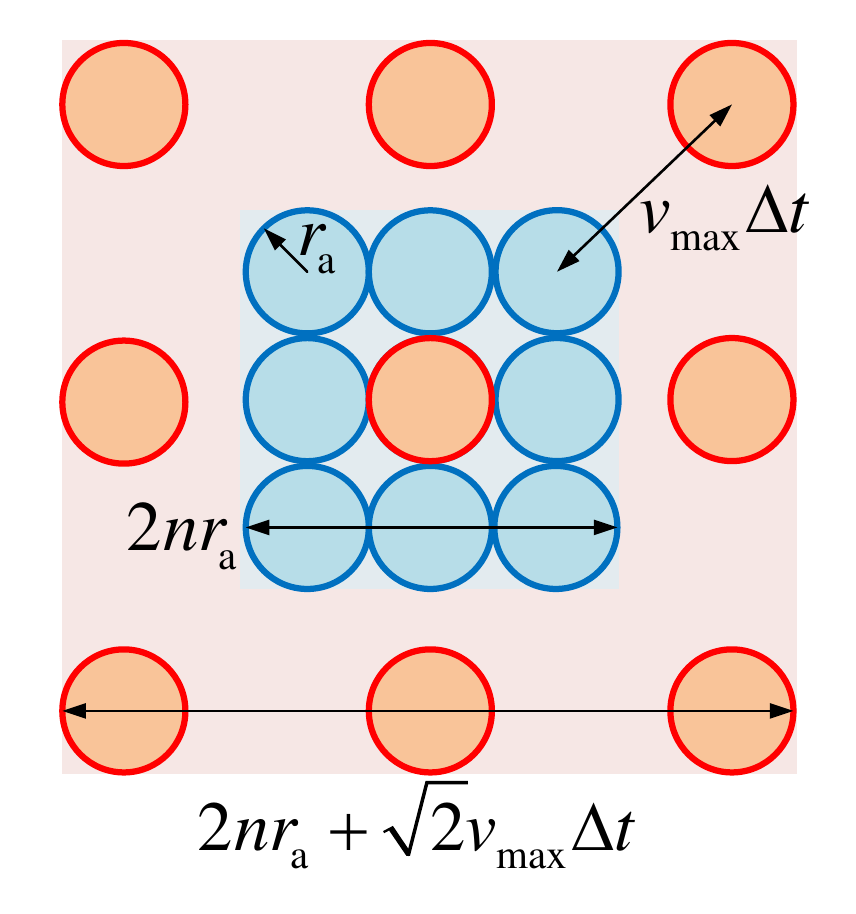}
	\caption{Schematic diagram of calculating the maximum change rate of the swarm density}
	\label{fig:appendix_new}
	\vspace{-0.6cm}
\end{figure}

\textit{\textbf{Assumption 4.}}
Suppose the area occupied by a robot swarm is rectangular. As shown in Fig. \ref{fig:appendix_new}, the fastest expansion strategy of the swarm is that the robots at the four corners of the original square area (the blue square) occupied by the swarm move away from the center point of the square with the maximum speed ${{v}_\text{max }}$, and then become the four corners of the new square area (the red square) occupied by the swarm.

\begin{itemize}
	\item \textbf{Speed Planning.} In order to accelerate the swarm through the virtual tube, speed planning is carried out. Based on \textit{Assumption 1}, we plan to obtain the average forward speed $ v_\text{a}^{*}(l)$ of the swarm at each position of the virtual tube. 
\end{itemize}

\begin{itemize}
	\item \textbf{Density Planning.} In order to ensure the safety of speed-constrained swarms within the virtual tube with varying widths, density planning is carried out. Based on \textit{Assumptions 2-4}, we obtain the swarm density ${{\rho }_\text{a}^{*}}(l)$ at each position of the virtual tube by planning.
\end{itemize}

\begin{itemize}
	\item \textbf{Tracking Control.} Let the swarm track the planning results of speed and density in the actual passing-through process. The planned average forward speed ${{v}_\text{a}^{*}}(l)$  is directly used as the forward speed component ${{v}_{\text{f},i}}$ in the controller to implement speed tracking. In addition, the avoidance radius ${{{r}}_{\text{a},i}}$ is under control to make the velocity command component ${{\mathbf{v}}_{\text{m},i}}$ and ${{\mathbf{v}}_{\text{xy},i}}$ changed, and then the planned swarm density ${{\rho }_\text{a}^{*}}(l)$ is tracked.
\end{itemize}


\section{MAIN RESULTS}

\subsection{Speed and Density Planning}

The goal is to plan the average forward speed ${{v}_\text{a}}(l)$ and density ${{\rho }_\text{a}}(l)$ of the swarm at each position of the virtual tube. In order to ensure the safety of the swarm, the density ${{\rho }_\text{a}}(l)$ is planned to be as close to the desired density ${{\rho }_\text{d}}$ as possible during the whole passing-through process. The constant ${{\rho }_\text{d}}$ is preset from experience as a reasonable value to ensure that the swarm passes through the virtual tube without collisions. Additionally, the average forward speed ${{v}_\text{a}}(l)$ is planned so that the swarm can pass through as fast as possible.


A pre-designed virtual tube is given.
According to the analysis above, the following planning is derived. The objective function and constraints are as follows:
\begin{equation}
	\setlength\abovedisplayskip{3pt}
	\setlength\belowdisplayskip{3pt}
	\underset{{{{v}_\text{a}}},{{{\rho}_\text{a}}}\in C\left[ 0,L \right]}{\mathop{ \min }}\,J=\int_{0}^{L}{\frac{1}{{{v}_\text{a}}\left( l \right)}\text{d}l}+\int_{0}^{L}{{{\left( {{\rho }_\text{a}}\left( l \right)-{{\rho }_\text{d}} \right)}^{2}}}\text{d}l
	\label{p1}
\end{equation}
subject to
\begin{equation}
	\setlength\abovedisplayskip{3pt}
	\setlength\belowdisplayskip{3pt}
	{{v}_\text{min }}\le {{v}_\text{a}}\left( l \right)\le {{v}_\text{max }},
	\label{p2}
\end{equation}
\begin{equation}
	\setlength\abovedisplayskip{3pt}
	\setlength\belowdisplayskip{3pt}
	\left| {{{\dot{v}}}_\text{a}} \right| \le {{a}_\text{v}},
	\label{p3}
\end{equation}
\begin{equation}
	\setlength\abovedisplayskip{3pt}
	\setlength\belowdisplayskip{3pt}
	{{v}_\text{a}}\left( l \right)\le \sqrt{{{a}_\text{n}}{{r}_{\text{t} }}\left( l \right)},
	\label{p4}
\end{equation}
\begin{equation}
	\setlength\abovedisplayskip{3pt}
	\setlength\belowdisplayskip{3pt}
	0< {{\rho }_\text{a}}\left( l \right)\le {{\rho }_\text{max }},
	\label{p5}
\end{equation}
\begin{equation}
	\setlength\abovedisplayskip{3pt}
	\setlength\belowdisplayskip{3pt}
	\left| {{{\dot{\rho }}}_\text{a}} \right| \le {{a}_{\rho }},
	\label{p6}
\end{equation}
\begin{equation}
	\setlength\abovedisplayskip{3pt}
	\setlength\belowdisplayskip{3pt}
	\begin{aligned}
		\left| \frac{{{\rho }_\text{f}}\left( l+\Delta l \right)-{{\rho }_\text{a}}\left( l \right)}{\Delta t} \right| \le {{a}_{\rho }}, \\ 
	\end{aligned}
	\label{p7}
\end{equation}
\begin{equation}
	\setlength\abovedisplayskip{3pt}
	\setlength\belowdisplayskip{3pt}
	{{a}_{\rho }}=-\frac{\sqrt{2}N{{v}_\text{max }}}{4{{n}^{3}}{{r}_{\text{a}}}^{3}}.
	\label{p8}
\end{equation}
Here, $L$, ${{v}_\text{min }}$, ${{v}_\text{max }}$, ${{a}_\text{v}}$, ${{a}_\text{n}}$, ${{\rho }_\text{max }}$, $N$ are known constants. The variable ${{r}_{\text{t} }}\left( l \right)$ denotes the radius of the curvature at the position with the center curve's arc length $l$. The constant  ${{\rho }_\text{max }}$ denotes the maximum density allowed for the swarm without colliding. The constant ${{a}_{\rho }}$ denotes the maximum change rate of the swarm density. The variable ${{\rho }_\text{f}\left( l + \Delta l \right)}$ denotes the predicted density when the swarm moves forward along the center curve without relative position change of any pair of robots based on ${{\rho}_\text{a}}\left( l \right)$. The constant $n=\left\lceil \sqrt{N} \right\rceil $, 
which denotes rounding up to the closest interger of $\sqrt{N}$.

$\bullet$ Equation (\ref{p1}) is the objective function. The first term represents the total time for the swarm to pass through the virtual tube. Therefore, the time for the swarm to pass through the whole virtual tube is shortened as much as possible when the first term is minimized. Meanwhile, the swarm density is close to the desired density ${{\rho }_\text{d}}$ at each position when the second term is minimized.

$\bullet$ Constraint  (\ref{p2}) limits the magnitude of the average forward speed ${{v}_\text{a}}(l)$, which is determined by the physical characteristics of robots according to Equation (\ref{v0}).

$\bullet$ Constraint  (\ref{p3}) limits the change rate of the average forward speed ${{v}_\text{a}}(l)$. The average forward speed ${{v}_\text{a}}(l)$ cannot be changed instantaneously, which is determined by the physical characteristics of robots according to Equation (\ref{v1}). Specifically,
\begin{equation*}
	\setlength\abovedisplayskip{3pt}
	\setlength\belowdisplayskip{3pt}
	\left| {{{\dot{v}}}_\text{a}} \right|\!=\left| \frac{\text{d}{{v}_\text{a}}\left( l \right)}{\text{d}t} \right|=\left| \frac{\text{d}{{v}_\text{a}}\left( l \right)}{\text{d}l}\cdot \frac{\text{d}l}{\text{d}t} \right|=\left| \frac{\text{d}{{v}_\text{a}}\left( l \right)}{\text{d}l} {{v}_\text{a}}\left( l \right) \right|\le {{a}_{v}}.
\end{equation*}

$\bullet$ Constraint (\ref{p4}) limits the magnitude of the average forward speed ${{v}_\text{a}}(l)$ at various locations of the virtual tube, which can be derived from Equation (\ref{v2}). Specifically, we need to ensure the swarm do not exceed the boundary of the virtual tube when it passes through the locations where the tube center curve is more curved. In other words, if the curvature of the tube center curve is large, ${{v}_\text{a}}(l)$ cannot be too large according to the speed constraint of robots in Equation (\ref{v2}).

$\bullet$ Constraint  (\ref{p5}) limits that the swarm density ${{\rho }_\text{a}}(l)$ cannot be greater than the maximum density ${{\rho }_\text{max }}$ for safety. Based on \textit{Assumption 2}, the minimum area occupied by the swarm is ${{S}_\text{min }}=N{r_{\text{p}}^{2}}$, thus the maximum density ${{\rho }_\text{max }}$ is 
\begin{equation*}
	\setlength\abovedisplayskip{3pt}
	\setlength\belowdisplayskip{3pt}
	{{\rho }_\text{max }}={N}/{{{S}_\text{min }}}={1}/{r_{\text{p}}^{2}}.
\end{equation*}

$\bullet$ Constraint (\ref{p6}) limits the change rate of the swarm density ${{\rho }_\text{a}}(l)$, that is, the area occupied by the swarm in the virtual tube cannot be changed instantly. The maximum change rate of the swarm density ${{a}_{\rho}}$ is calculated by Equation (\ref{p8}). Specifically,
\begin{equation*}
	\setlength\abovedisplayskip{3pt}
	\setlength\belowdisplayskip{3pt}
	\left| {{{\dot{\rho }}}_\text{a}} \right|=\left| \frac{\text{d}{{\rho }_\text{a}}\left( l \right)}{\text{d}t} \right|=\left| \frac{\text{d}{{\rho }_\text{a}}\left( l \right)}{\text{d}l}\cdot \frac{\text{d}l}{\text{d}t} \right|=\left| \frac{\text{d}{{\rho }_\text{a}}\left( l \right)}{\text{d}l} {{v}_\text{a}}\left( l \right) \right|\le {{a}_{\rho }}.
\end{equation*}



$\bullet$ Constraint (\ref{p7}) provides predictive density planning according to the pre-designed virtual tube with known parameters. As shown in Fig. \ref{fig:pf}, ${{\rho }_\text{f}}\left( l+\Delta l \right)$ represents the predicted swarm density after traversing a distance $\Delta l $ along the tube center curve without changing the relative position of any pair of robots from the position with arc length $l$ of the tube center curve, where the swarm density is ${{\rho }_\text{a}}(l)$. The variable ${{\rho }_\text{f}}\left( l+\Delta l \right)$ can be calculated according to Equations (\ref{p0}) and (\ref{ps}), which is related to $N$ and $\lambda \left( l \right)$. Thus, according to Equation (\ref{p7}),
\begin{equation*}
	\begin{aligned}
		\setlength\abovedisplayskip{3pt}
		\setlength\belowdisplayskip{3pt}
		& \left| \frac{{{\rho }_\text{f}}\left( l+\Delta l \right)-{{\rho }_\text{a}}\left( l \right)}{\Delta t} \right|=\left| \frac{{{\rho }_\text{f}}\left( l+\Delta l \right)-{{\rho }_\text{a}}\left( l \right)}{\Delta l}{{v}_\text{a}}\left( l \right) \right| \\ 
		& =f\left( N,\lambda \left( l \right),{{\rho }_\text{a}}\left( l \right),{{v}_\text{a}}\left( l \right),\Delta l \right)\le {{a}_{\rho }}, \\ 
	\end{aligned}
\end{equation*}
where $f\left( N,\lambda \left( l \right),{{\rho }_\text{a}}\left( l \right),{{v}_\text{a}}\left( l \right),\Delta l \right)$ denotes a function related to $ N,\lambda \left( l \right),{{\rho }_\text{a}}\left( l \right),{{v}_\text{a}}\left( l \right),\Delta l $. 
Based on \textit{Assumption 3}, this formula means that if the swarm moves forward without changing the relative position of any pair of robots, the change rate of the swarm density caused by the variation of the tube width cannot exceed the maximum change rate of the swarm density ${{a}_{\rho}}$ calculated by Equation (\ref{p8}).
It can be found from the following planning results that this constraint plans a small swarm density before entering the narrowest part of the virtual tube. Thus, the swarm expands before entering the narrowest part of the virtual tube. In other words, the swarm can compensate the increase of density caused by the varying tube width through the active expansion in advance. In conclusion, conflict and congestion are avoided before the swarm enters the narrowed part of the virtual tube according to Equation (\ref{p7}). Therefore, safety is ensured, and efficiency is improved. 


$\bullet$ Constraint (\ref{p8}) indicates that the maximum change rate of the swarm density ${{a}_{\rho }}$ relies on $N,{{v}_{\max }},{{r}_{\text{a}}}$. Intuitively, the larger the maximum speed ${{v}_{\max }}$ is, meaning that the swarm can expand faster, thus the larger ${{a}_{\rho }}$ can be. Based on the fastest expansion strategy defined by \textit{Assumption 4}, detailed derivations are shown as follows. 

The maximum change rate of the swarm density ${{a}_{\rho }}$ is derived as follows.
Based on \textit{Assumption 4}, swarm density changes most rapidly when the swarm expands fastest. As shown in Fig. \ref{fig:appendix_new}, assume that the robots are located within the blue square initially, and the side length of the blue square is $2n{{r}_{\text{a}}}$. Assuming that the density changes most rapidly, after time $\Delta t$, the robots expand to the red square in Fig. \ref{fig:appendix_new}, and the length of the red square is $2n{{r}_{\text{a}}}+\sqrt{2}{{v}_\text{max }}\Delta t$. Therefore, the maximum change rate of the swarm density is
\begin{equation*}
	\setlength\abovedisplayskip{3pt}
	\setlength\belowdisplayskip{3pt}
	{{a}_{\rho }}=\underset{\Delta t\to 0}{\mathop{\lim }}\,\frac{{{\rho }_{1}}-{{\rho }_{0}}}{\Delta t}=\underset{\Delta t\to 0}{\mathop{\lim }}\,\frac{\frac{N}{{{S}_{1}}}-\frac{N}{{{S}_{0}}}}{\Delta t}=-\frac{\sqrt{2}N{{v}_\text{max }}}{4{{n}^{3}}{{r}_{\text{a}}}^{3}},
\end{equation*}
where ${{\rho }_{0}}$ represents the initial swarm density, and ${{\rho }_{1}}$ represents the swarm density after time $\Delta t$.

\subsection{Tracking Control of Planned Average Forward Speed and Density}

The purpose of control is to make the swarm follow the planned average forward speed  ${{v}_\text{a}^{*}}(l)$ and swarm density ${{\rho }_\text{a}^{*}}(l)$ during the passing-through process.

\subsubsection{Track Planned Average Forward Speed}
In order to make the real-time average forward speed of the swarm track the planned average forward speed $v_\text{a}^{*}\left( l \right)$, the planned average speed $v_\text{a}^{*}\left( l \right)$ is directly used as the forward speed component ${{v}_{\text{f},i}}$ of the $i$th robot based on the robot controller (\ref{a1}) as follows:
\begin{equation*}
	\setlength\abovedisplayskip{3pt}
	\setlength\belowdisplayskip{3pt}
	{{v}_{\text{f},i}}\left( {{l}_{i}} \right) = v_\text{a}^{*}\left( {{l}_{i}} \right), {l}_{i}\in \left[ 0,L \right].
\end{equation*}

\begin{figure}[!t]\centering
	\centering
	\setlength{\abovecaptionskip}{-0.2cm}
	\includegraphics[width=0.8\linewidth]{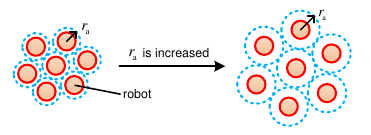}
	\caption{Swarm density will be decreased if ${{{r}}_{\text{a}}}$ is increased.}
	\label{f_density}
	\vspace{-0.5cm}
\end{figure}

\subsubsection{Track Planned Swarm Density}
Density tracking is realized by changing the avoidance radius ${{{r}}_{\text{a}}}$.
So as to make the real-time swarm density ${{\rho }_\text{r}}(l)$ follow the planned swarm density ${{\rho }_\text{a}^{*}}(l)$ during the passing-through process, the speed control component ${{\mathbf{v}}_{\text{m},i}}$ and ${{\mathbf{v}}_{\text{xy},i}}$ of each robot are changed by different setting of ${{{r}}_{\text{a},i}}$, 
which is a variable. Then the area occupied by the swarm within the virtual tube is changed. Finally, the real-time swarm density ${{\rho }_\text{r}}(l)$ is controlled to be close to the planned swarm density ${{\rho }_\text{a}^{*}}(l)$ as much as possible. Therefore, the controller for ${{{r}}_{\text{a},i}}$ of the $i$th robot is designed as follows:
\begin{equation}
	\setlength\abovedisplayskip{3pt}
	\setlength\belowdisplayskip{3pt}
	{{r}_{\text{ac},i}}({l}_{i})=\left\{ \begin{matrix}
		{0} & {{\rho }_\text{r}}({l}_{i})\le {{\rho }_\text{a}^{*}}\left( {l}_{i} \right)  \\
		{{k}_{{{r}_{\text{a}}}}}\left( {{\rho }_\text{r}}({l}_{i})-{{\rho }_\text{a}^{*}}\left( {l}_{i} \right) \right) & {{\rho }_\text{r}}({l}_{i})>{{\rho }_\text{a}^{*}}\left( {l}_{i} \right)  \\
	\end{matrix} \right.	,
	\label{a3}
\end{equation}
where ${{k}_{{{r}_{\text{a}}}}}>0$ is a coefficient, ${l}_{i}\in \left[ 0,L \right]$. According to the law (\ref{a3}), when the real-time swarm density ${{\rho }_\text{r}}({l}_{i})$ is larger than the planned swarm density ${{\rho }_\text{a}^{*}}({l}_{i})$, the avoidance radius ${{{r}}_{\text{a},i}}$ will be increased. Therefore, as shown in Fig. \ref{f_density}, the real-time swarm density ${{\rho }_\text{r}}({l}_{i})$ will be decreased to follow the planned swarm density ${{\rho }_\text{a}^{*}}({l}_{i})$, which brings an expansion that avoids collisions among robots to ensure the safety of the swarm passing-through process. 

Particularly, 
the density tracking is no longer considered when the real-time swarm density ${{\rho }_\text{r}}({l}_{i})$ is less than the planned swarm density ${{\rho }_\text{a}^{*}}({l}_{i})$ in Equation (\ref{a3}). The reason is that there are no safety risks for the swarm when the real-time swarm density ${{\rho }_\text{r}}({l}_{i})$ is less than the planned swarm density ${{\rho }_\text{a}^{*}}({l}_{i})$ according to Equation (\ref{p5}).


\section{SIMULATION AND EXPERIMENT RESULTS}
\subsection{Numerical Simulation}
\subsubsection{Simulation With Planning and Without Planning}

In the following simulation, the passing-through process of the swarm within the virtual tube is planned by the segmented planning method.
The average forward speed ${{v}_\text{a}}(l)$ and density ${{\rho }_\text{a}}(l)$ of the swarm to be planned  are represented in the form of a third-order polynomial as follows:
\begin{equation*}
	\setlength\abovedisplayskip{3pt}
	\setlength\belowdisplayskip{3pt}
	\begin{aligned}
		& {{v}_\text{a}}\left( l \right)={{c}_{3}}{{l}^{3}}+{{c}_{2}}{{l}^{2}}+{{c}_{1}}l+{{c}_{0}}, \\
		& {{\rho }_\text{a}}\left( l \right)={{b}_{3}}{{l}^{3}}+{{b}_{2}}{{l}^{2}}+{{b}_{1}}l+{{b}_{0}},\\
	\end{aligned}
\end{equation*}
where ${{c}_{3}}, {{c}_{2}}, {{c}_{1}}, {{c}_{0}}, {{b}_{3}}, {{b}_{2}},
{{b}_{1}}, {{b}_{0}}$ are the coefficients of the third-order polynomial, which are going to be determined by (\ref{p1})-(\ref{p8}).

\begin{figure}[!t]
	\centering
	\setlength{\abovecaptionskip}{-0.3cm}
	\includegraphics[width=0.8\linewidth]{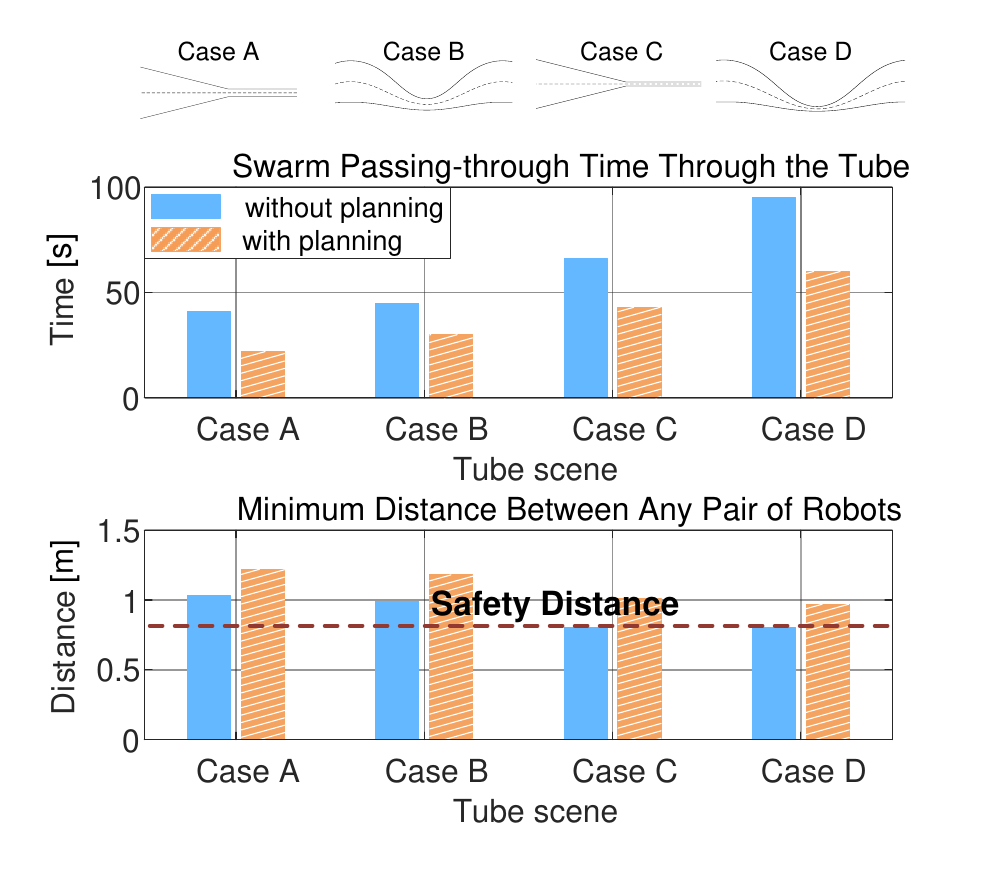}
	\caption{Swarm passing-through time and the minimum distance between any pair of robots in four different virtual tube scenes without planning and with planning.}
	\label{fig:f220909_4}
	\vspace{-0.3cm}
\end{figure}

Simulation comparisons between control without and with planning in various virtual tube scenes are carried out. Specifically, \emph{control without planning} refers to controlling with default parameters based on our previous control method \cite{b17}, which refers to Equation (\ref{s1}), while \emph{control with planning} refers to controlling according to the planned average forward speed $v_\text{a}^{*}\left( l \right)$ and density ${{\rho }_\text{a}^{*}}(l)$ of the swarm. \emph{Virtual tube scenes} include a normally narrowing trapezoidal virtual tube, a normally narrowing curved virtual tube, a rapidly narrowing trapezoidal virtual tube, and a rapidly narrowing curved virtual tube, corresponding to the case A, B, C, and D in Fig. \ref{fig:f220909_4} respectively. Parameter settings are shown in Table \ref{tab:1}. In addition, the \emph{passing-through time} is the assessment for \emph{efficiency}, and the \emph{minimum distance} between any pair of robots during the passing-through process is the assessment for \emph{safety}.

\begin{table}[!t]
	\centering
	\caption{Parameter Settings}
	\label{tab:1}
	\renewcommand\arraystretch{0.8}
	\begin{tabular}{cccccc }
		\hline\hline\noalign{\smallskip}	
		Parameter & \emph{N} & ${{v}_\text{min }}$ & ${{v}_\text{max }}$& ${{a}_\text{v}}$& ${{a}_\text{n}}$ \\
		\noalign{\smallskip}\hline\noalign{\smallskip}
		Value & 20 & 2 &5 &1 &1  \\
		\noalign{\smallskip}\hline\noalign{\smallskip}
		Parameter &${{r}_{\text{p}}}$ & ${{r}_{\text{s}}}$ &${{r}_{\text{a}}}$ &${{\rho }_\text{d}}$ &${{\rho }_\text{max }}$\\
		\noalign{\smallskip}\hline\noalign{\smallskip}
		Value  &0.3 &0.4 &0.8 &0.1989 &0.9974 \\
		\noalign{\smallskip}\hline\hline\noalign{\smallskip}
	\end{tabular}
	\vspace{-0.4cm}
\end{table}




It can be observed from Fig. \ref{fig:f220909_4} that the passing-through time of the swarm with planning is much smaller than without planning. Additionally, the minimum distance between any pair of robots with planning is larger than that without planning, which clearly shows that speed and density planning improves the efficiency and ensures the safety of the swarm passing-through process effectively.

\begin{figure}[!t]
	\centering
	\setlength{\abovecaptionskip}{-0.3cm}
	\includegraphics[width=0.9\linewidth]{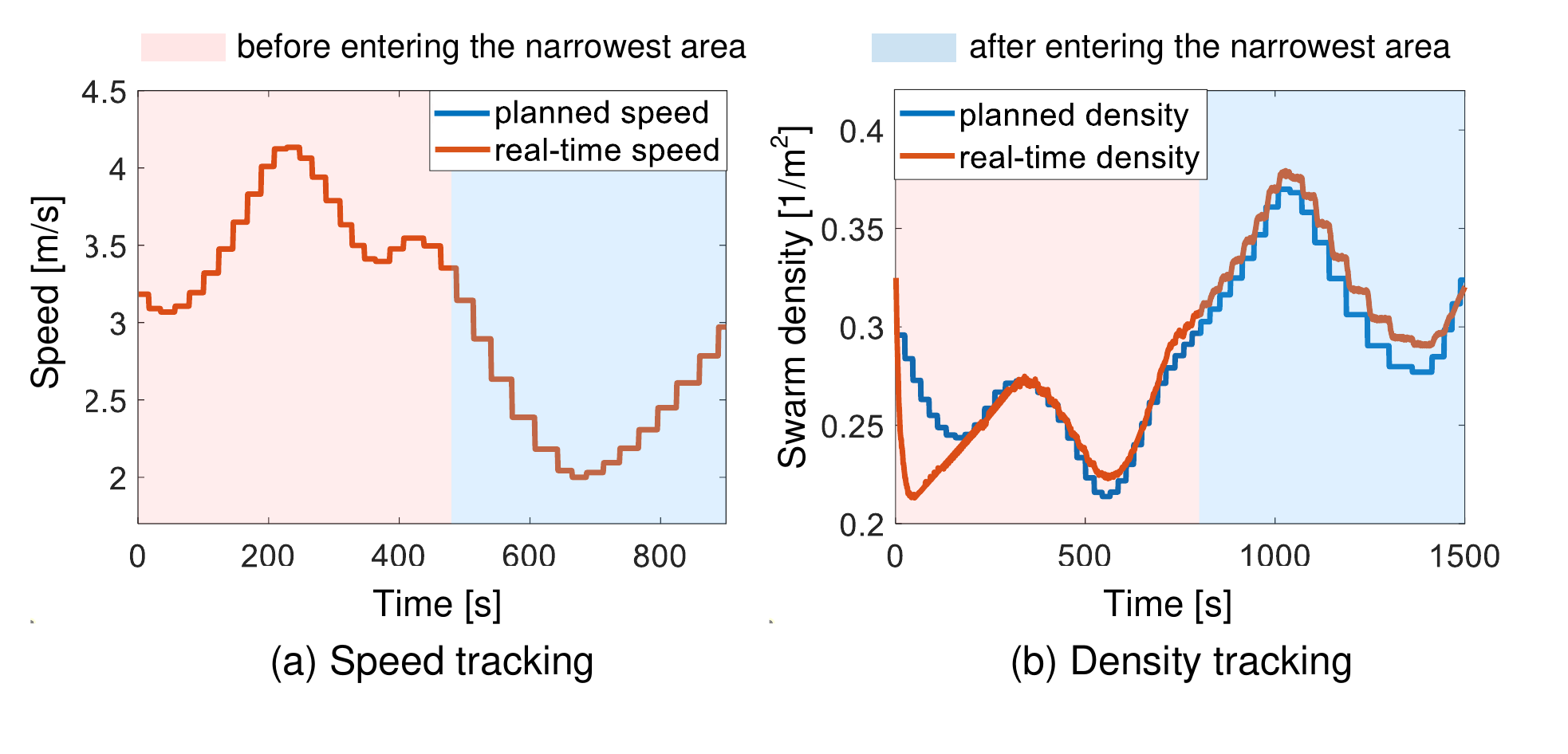}
	\caption{Speed tracking and density tracking in a normally narrowing trapezoidal virtual tube.}
	\label{fig:f220909_3}
	\vspace{-0.5cm}
\end{figure}



\begin{figure}[!t]
	\centering
	\setlength{\abovecaptionskip}{-0.6cm}
	\includegraphics[width=1\linewidth]{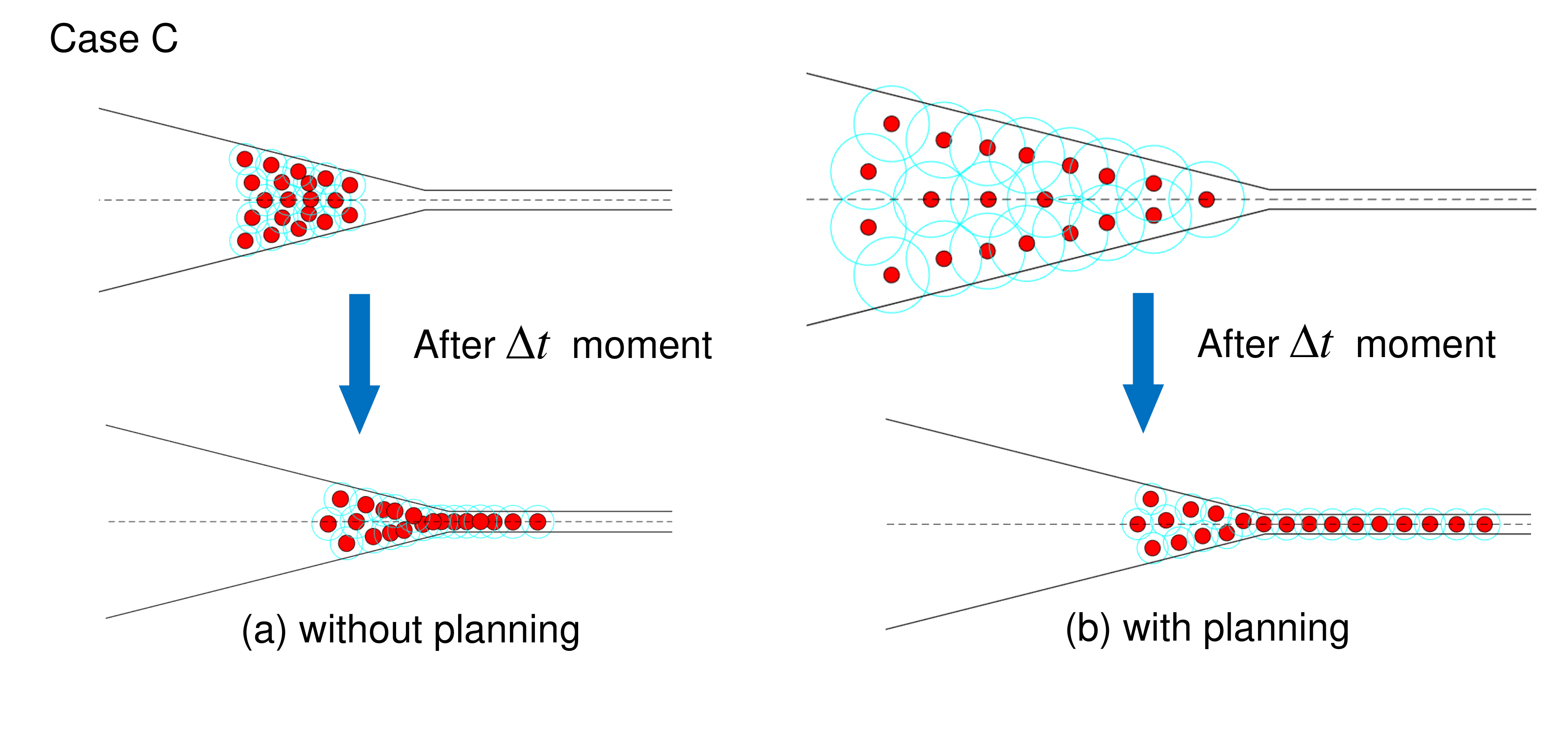}
	\caption{Comparison of swarm passing through a rapidly narrowing trapezoidal virtual tube without planning and with planning.}
	\label{fig:f220909_1}
	\vspace{-0.3cm}
\end{figure}
\begin{figure}[!t]
	\centering
	\setlength{\abovecaptionskip}{-0.6cm}
	\includegraphics[width=1\linewidth]{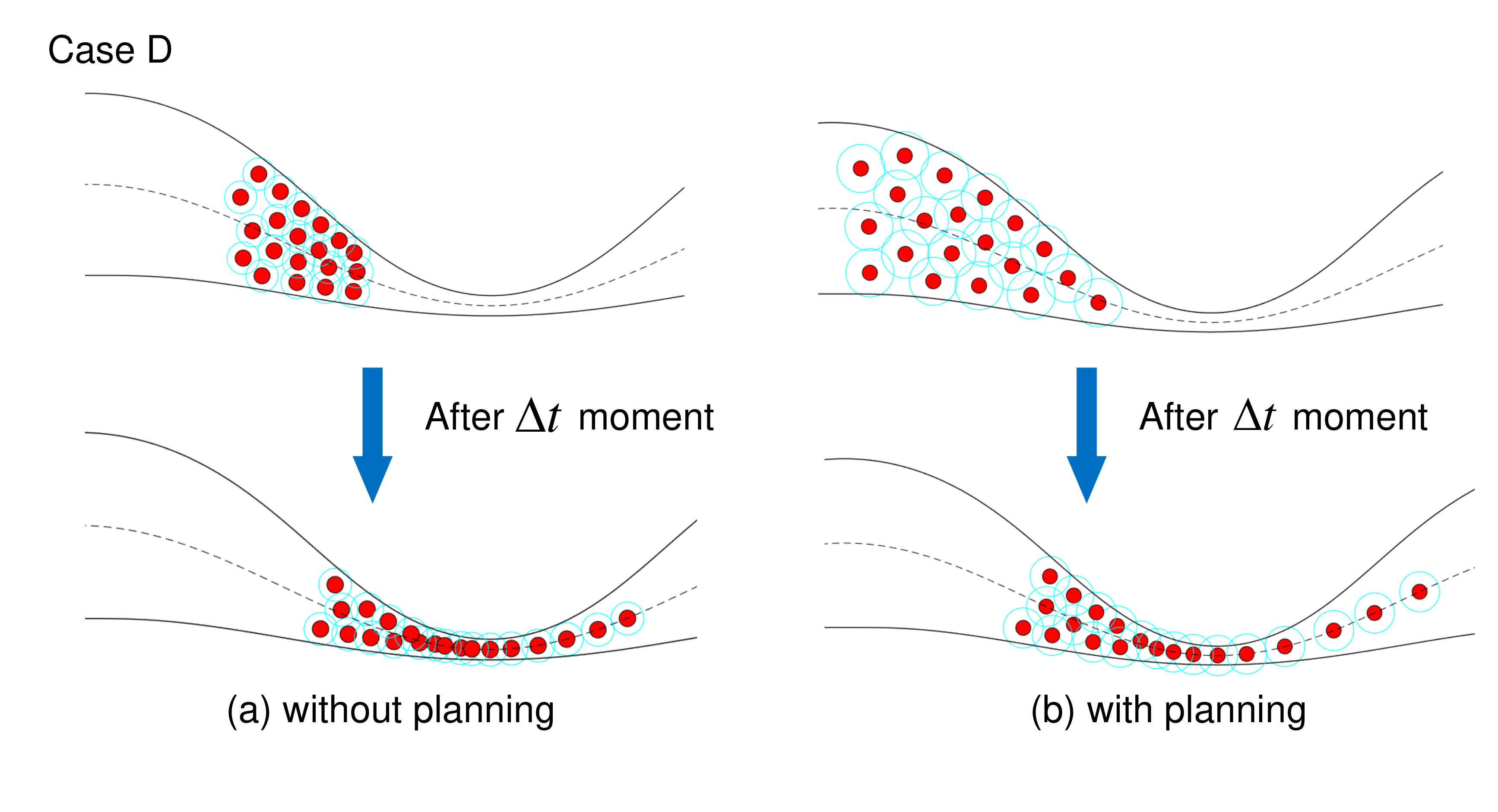}
	\caption{Comparison of swarm passing through a rapidly narrowing curved virtual tube without planning and with planning.}
	\label{fig:f220909_2}
	\vspace{-0.3cm}
\end{figure}

It is easy to observe from Fig. \ref{fig:f220909_3} that the swarm can be controlled to track the planning results of speed and density very well. Particularly, because the real-time swarm density ${{\rho }_\text{r}}(l)$ is always larger than the planned swarm density ${{\rho }_\text{a}^{*}}(l)$ in Fig. \ref{fig:f220909_3}, the density tracking is well implemented according to Equation (\ref{a3}). Moreover, it can be found in Fig. \ref{fig:f220909_3} that there is a significant decrease in the planned density before entering the narrowest part of the virtual tube, which indicates that the planned density requires the swarm to expand before entering the narrowest part, and it is consistent with the simulation results. 
Additionally, it can be observed from Fig. \ref{fig:f220909_1} and Fig. \ref{fig:f220909_2} that there are collisions among robots before entering the narrowest part of the virtual tube without planning. However, due to the expansion of the swarm in advance, the collisions are avoided with planning. 

In conclusion, the method proposed in this paper is suitable for various virtual tube scenes. In addition, the appropriate average forward speed $v_\text{a}^{*}\left( l \right)$ and density ${{\rho }_\text{a}^{*}}(l)$ of the swarm are planned, and the controller implements the real-time tracking of the planning results, which increases the efficiency and ensures the safety of the swarm passing-through process in various virtual tubes of varying widths significantly.

\subsubsection{Comparative Simulation}
\begin{figure}[!t]
	\centering
	\setlength{\abovecaptionskip}{-0.5cm}
	\includegraphics[width=0.8\linewidth]{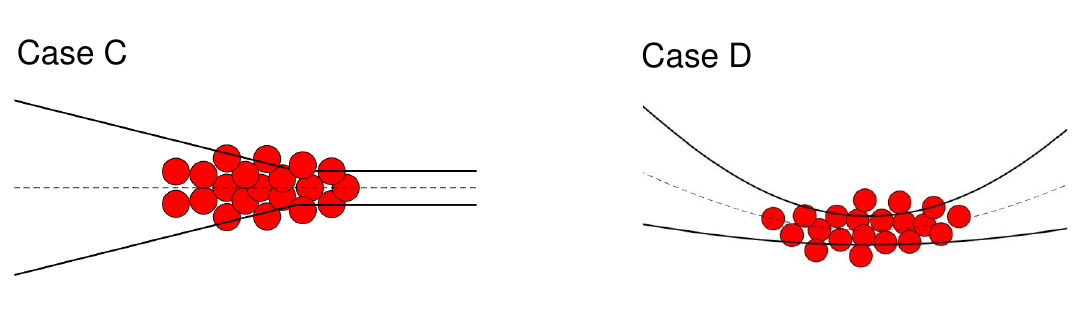} 
	\caption{Distribution of the swarm when passing through the narrowest part of the rapidly narrowing trapezoidal virtual tube and rapidly narrowing curved virtual tube under the method in paper \cite{b18}.}
	\label{fig:Flast}
	\vspace{-0.3cm}
\end{figure}

a) Compare with the optimized flocking method \cite{b18}. Simulations based on the optimized flocking method as well as the method proposed in this paper are performed in the same virtual tube scenes as follows.
The optimized flocking method is a control method to ensure that large flocks of autonomous drones seamlessly navigate in confined spaces, which has been widely used recently. It can be observed from Fig. \ref{fig:Flast} that the robots collide with each other around the narrowest part even if the minimum speed is $0$. 

\begin{figure}[!t]
	\centering
	\setlength{\abovecaptionskip}{-0.4cm}
	\includegraphics[width=1\linewidth]{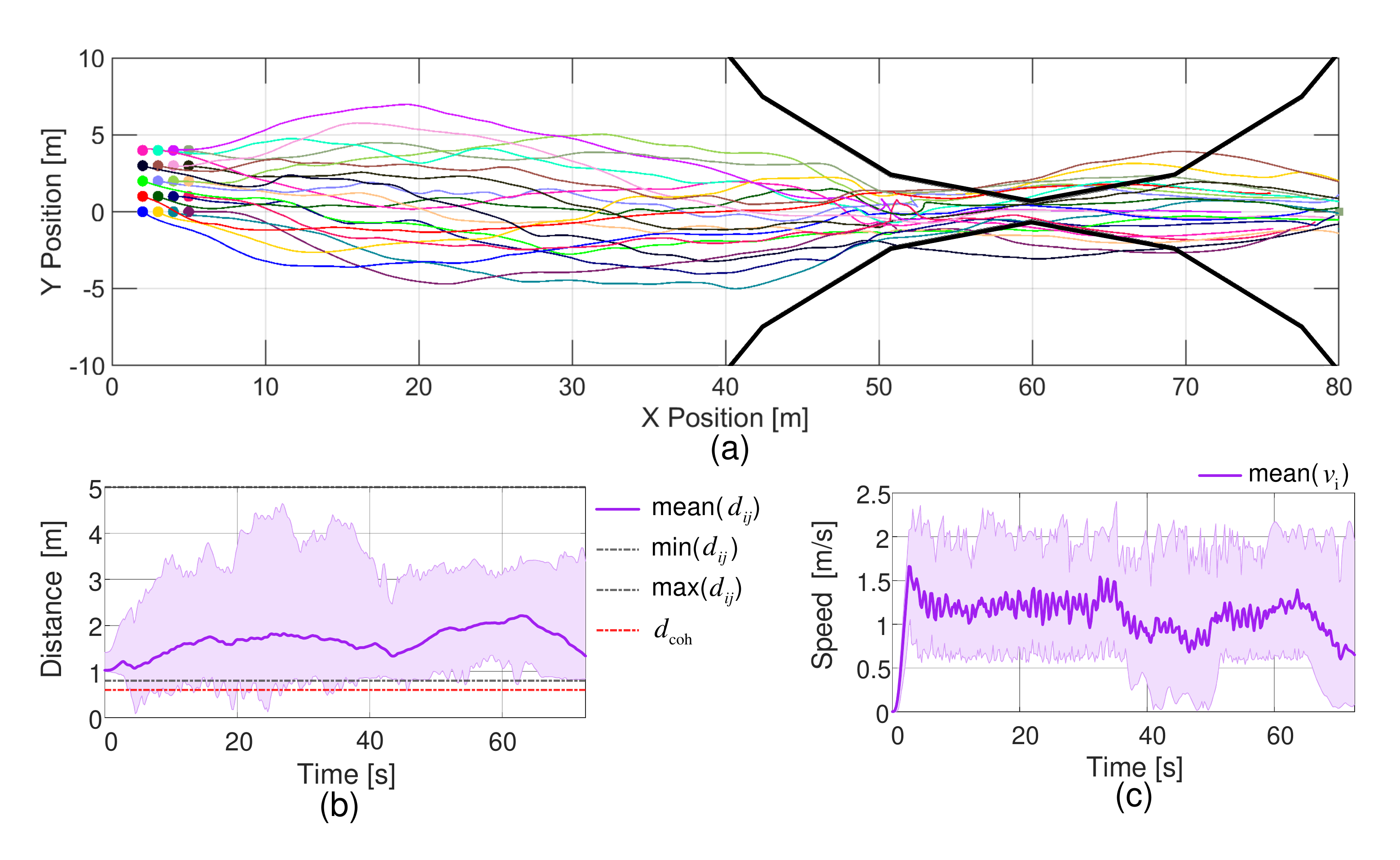} 
	\caption{The swarm composed of twenty robots passes through a narrowing curved virtual tube based on NMPC. (a) The planned trajectories of the robots in the swarm. (b) Inter-agent distance average (solid purple line), range (shaded region), cohesion distance (upper grey dotted line), safety distance (lower grey dotted line) and collision distance (red dotted line). (c) Swarm speed average (solid line) and range (shaded region).}
	\label{fig_3d}
	\vspace{-0.5cm}
\end{figure}



b) Compare with the nonlinear model predictive control (NMPC) \cite{b24}. The NMPC method is an effective method for aerial swarms to pass through cluttered environments. We establish a similar narrowing curved virtual tube as Case D. Specifically, the width of the narrowest part of this tube is the same as Case D. Moreover, The relevant parameters are set the same as shown in Table \ref{tab:1}. The minimum speed is set to 0. Then twenty robots are controlled to pass through the narrowing virtual tube based on NMPC. The planned trajectories are shown in Fig. \ref{fig_3d} (a). We can easily observe some excess of the tube boundaries before entering the narrowest area of the tube. Additionally, it can be observed from Fig. \ref{fig_3d} (b) that the minimum inter-agent distance is lower than the safety distance in the most of the time and lower than the collision distance sometimes, which represents some collisions between multiple robots are already occurred. Furthermore, the average speed is not fast according to Fig. \ref{fig_3d} (c). Therefore, it is impossible to use merely NMPC to make the swarm pass through a narrow space without collisions between robots and obstacles.

In conclusion, compared the above two simulations with the simulation in Fig. \ref{fig:f220909_1} and Fig. \ref{fig:f220909_2}, the method proposed in this paper can improve the safety and efficiency of the swarm's traversing process in narrow spaces to a large extent.

\subsection{Experiments}
%
%
%
%
Based on the method proposed in this paper, experiments are conducted on Robotarium \cite{b23} and different types of quadcopters as follows, which verifies its application on various experimental platforms.

\subsubsection{Experiments on Robotarium}
\begin{figure}[!t]
	\centering
	\setlength{\abovecaptionskip}{-0.2cm}
	\includegraphics[width=0.9\linewidth]{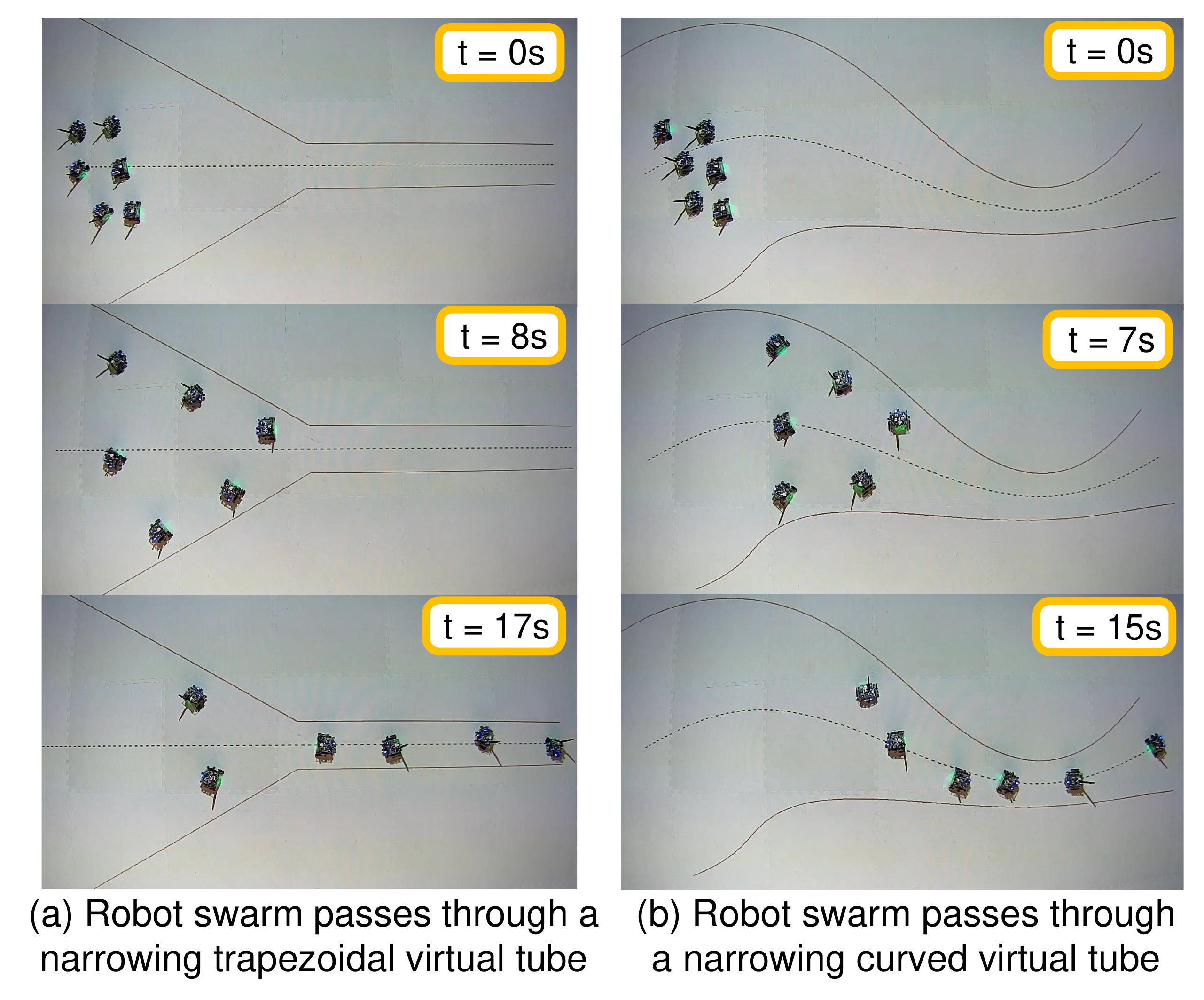}
	\caption{Experiments on Robotarium.}
	\label{car}
	\vspace{-0.4cm}
\end{figure}

A speed-constrained robot swarm consisting of six robots is used to do experiments on Robotarium. This swarm is required to pass through a narrowing trapezoidal virtual tube and a narrowing curved virtual tube. As shown in Fig. \ref{car}, based on the speed and density planning, there are apparent expansions before entering the narrowest part of both virtual tubes around $7$ seconds after departure. Therefore, the swarm passes through the both virtual tubes without conflict or going beyond the tube boundary finally. It can be inferred from these experiments that the safety of the passing-through process is ensured by the method proposed.

\subsubsection{Experiments on Quadcopters}
\begin{figure}[!t]
	\centering
	\setlength{\abovecaptionskip}{-0.1cm}
	\includegraphics[width=0.8\linewidth]{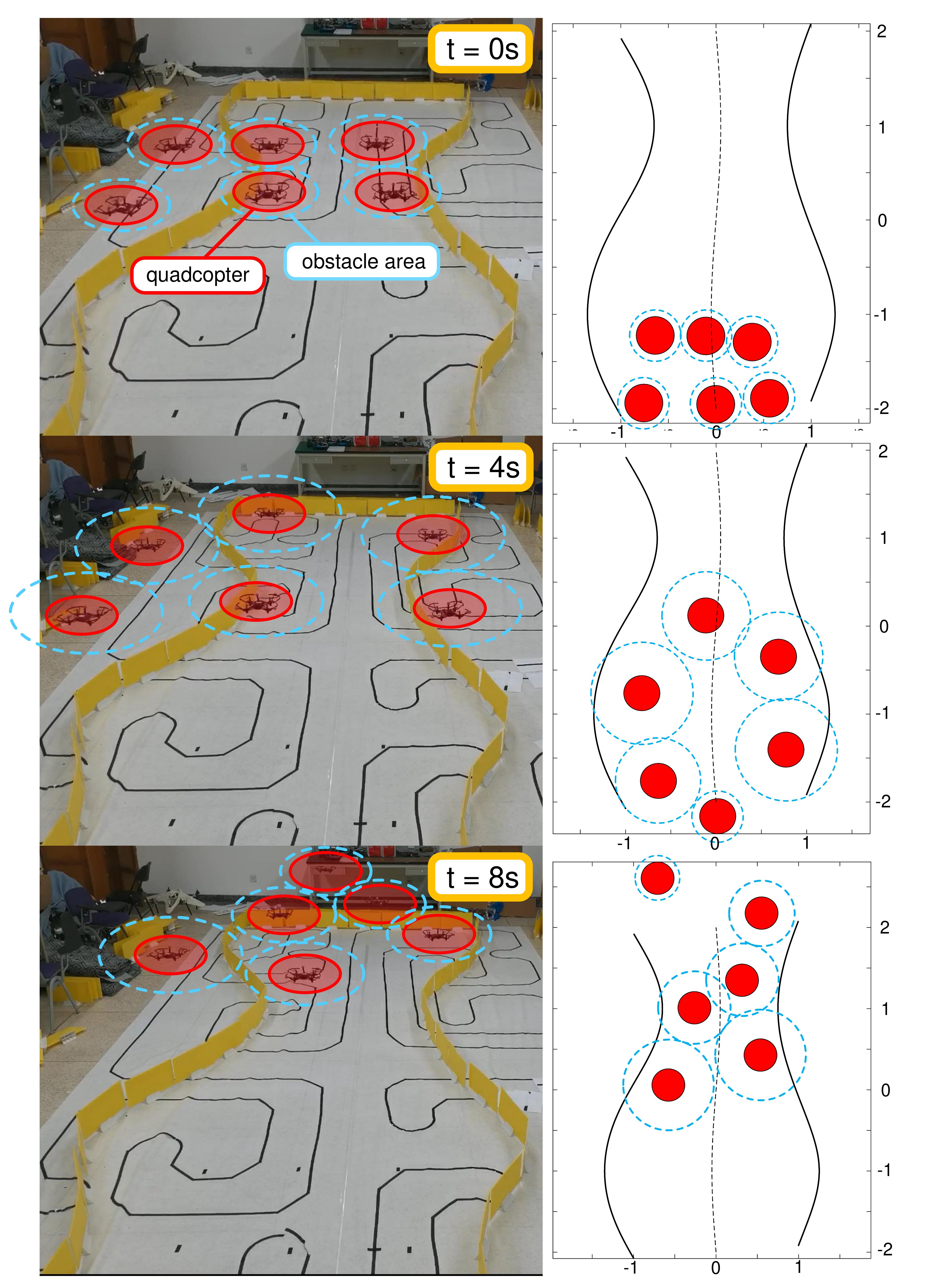} 
	\caption{Flight experiment on quadcopters in a virtual tube.}
	\label{fig:ff33}
	\vspace{-0.6cm}
\end{figure}

Six speed-constrained quadcopters are used to do experiments within a narrowing curved virtual tube and simulate in real time. As shown in Fig. \ref{fig:ff33}, the blue dotted line  denotes the avoidance radius ${{{r}}_{\text{a}}}$.  There is an obvious expansion of the quadcopter swarm before entering the narrowest part of the virtual tube at the $4$ second due to the increase of  ${{{r}}_{\text{a}}}$.  Finally, the quadcopter swarm passes through the narrowest part of the virtual tube without conflict or going beyond the tube boundary at the $8$ second. 

\begin{figure}[!t]
	\centering
	\setlength{\abovecaptionskip}{-0.3cm}
	\includegraphics[width=0.9\linewidth]{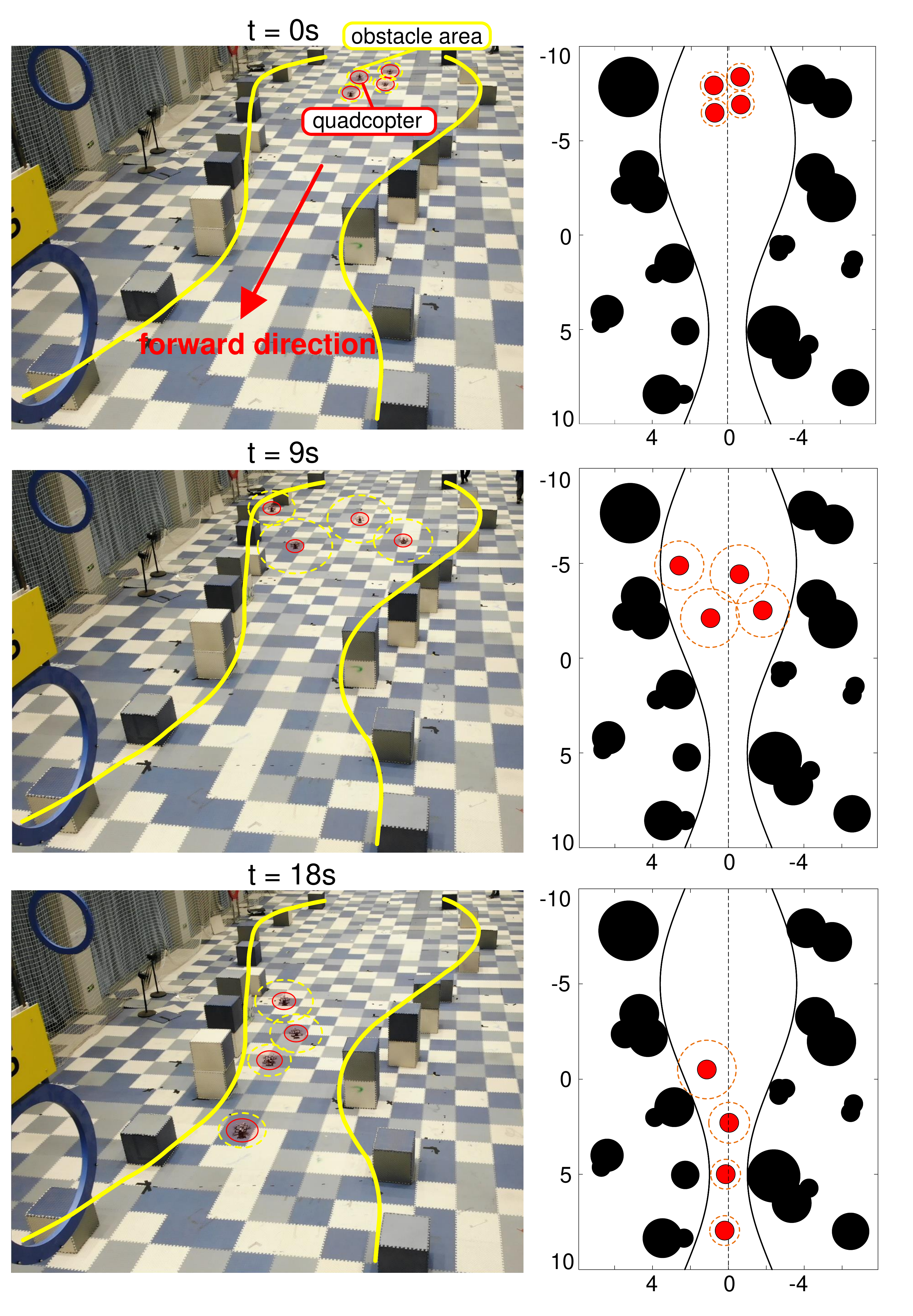} 
	\caption{Flight experiment on quadcopters with real obstacles.}
	\label{sh}
	\vspace{-0.3cm}
\end{figure}

\begin{figure}[!t]
	\centering
	\setlength{\abovecaptionskip}{-0.3cm}
	\includegraphics[width=0.8\linewidth]{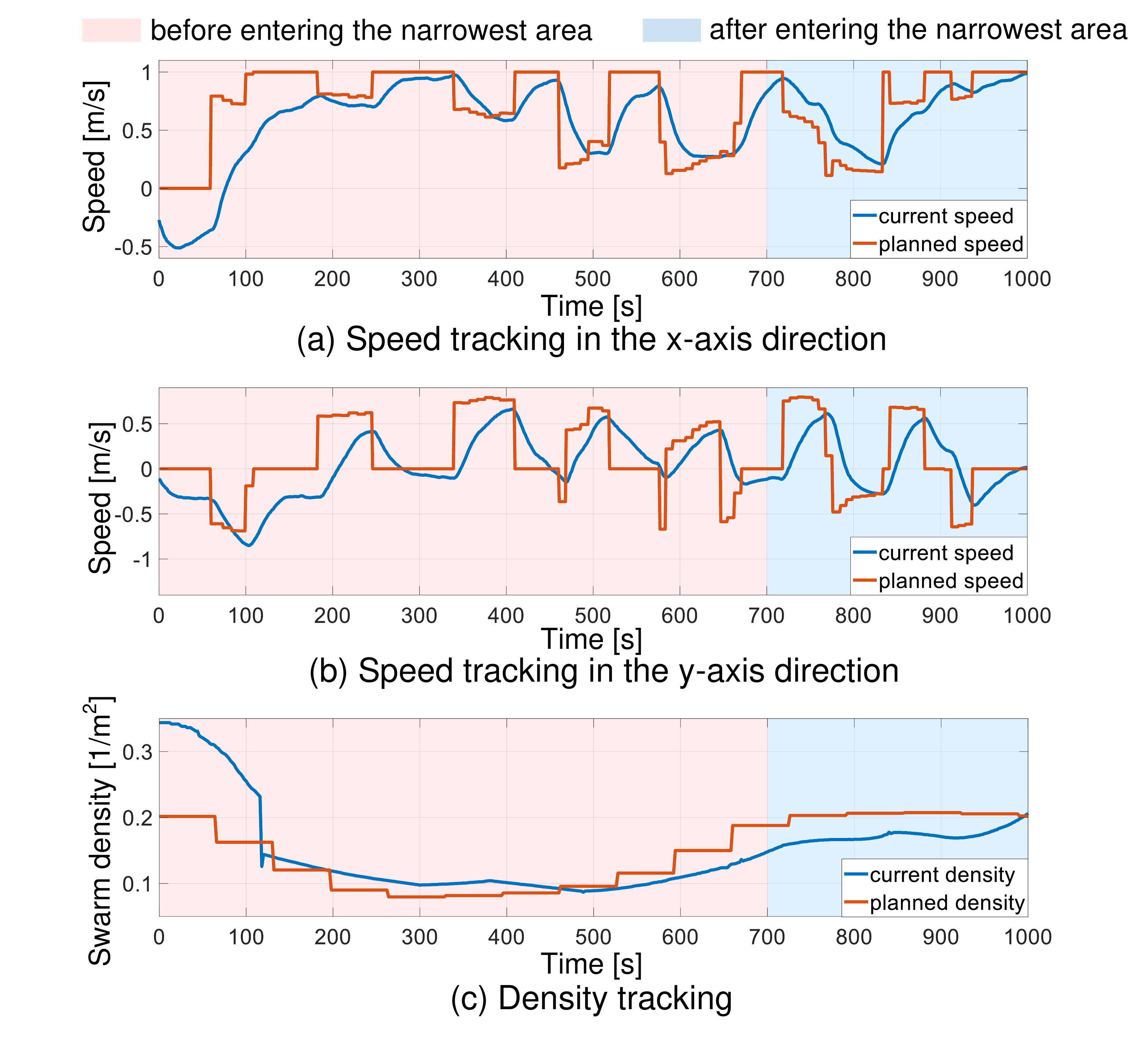} 
	\caption{Speed and density tracking of one quadcopter in the flight experiment with real obstacles.}
	\label{ex_track}
	\vspace{-0.7cm}
\end{figure}

An experiment based on another type of quadcopters is also carried out in a complex environment with real obstacles. In this experiment, the quadcopters rely on on-board computers to make decisions, achieving a truly distributed control. Specifically, we use optitrack motion capture to provide precise positions of quadcopters and obstacles, and use Jetson Xavier NX as the core board. As shown in Fig. \ref{sh}, based on an advanced expansion, each quadcopter of the swarm safely passes through the narrowest area where the obstacles are dense. The tracking curves of speed and density during the whole passing-through process are shown in Fig. \ref{ex_track}, which indicates that the swarm can be controlled to track the planned speed and density very well.

\section{CONCLUSIONS}
Speed and density planning with tracking control is proposed in this paper to solve the problem for a speed-constrained robot swarm passing through a known virtual tube with varying widths. The method proposed greatly improves the safety and efficiency of the swarm{'}s passing-through process. It has potential in air traffic of drones, a robot swarm passing through a tunnel, and a robot swarm searching in a cluttered environment, etc..
However, inaccurate robot tracking of the planned speed and density may cause collisions among robots sometimes, which is still deserved to study in the future. Influencing factors are as follows. (i) The narrowing degree of the virtual tube. (ii) The limitation of the swarm's ability to expand and track the control command. (iii) The inappropriate settings of the initial states and ideal states. (iv) The length of virtual tubes and the number of robots in the swarm.





\bibliographystyle{IEEEtran} 
\bibliography{root}

\begin{thebibliography}{10}
\providecommand{\url}[1]{#1}
\csname url@rmstyle\endcsname
\providecommand{\newblock}{\relax}
\providecommand{\bibinfo}[2]{#2}
\providecommand\BIBentrySTDinterwordspacing{\spaceskip=0pt\relax}
\providecommand\BIBentryALTinterwordstretchfactor{4}
\providecommand\BIBentryALTinterwordspacing{\spaceskip=\fontdimen2\font plus
\BIBentryALTinterwordstretchfactor\fontdimen3\font minus
  \fontdimen4\font\relax}
\providecommand\BIBforeignlanguage[2]{{%
\expandafter\ifx\csname l@#1\endcsname\relax
\typeout{** WARNING: IEEEtran.bst: No hyphenation pattern has been}%
\typeout{** loaded for the language `#1'. Using the pattern for}%
\typeout{** the default language instead.}%
\else
\language=\csname l@#1\endcsname
\fi
#2}}

\bibitem{b40}
Q.~Quan, \emph{{I}ntroduction to {M}ulticopter {D}esign and {C}ontrol}.\hskip
  1em plus 0.5em minus 0.4em\relax Springer, 2017.

\bibitem{b3}
Y.~Xu, S.~Zhao, D.~Luo, and Y.~You, ``Affine formation maneuver control of
  high-order multi-agent systems over directed networks,'' \emph{Automatica},
  vol. 118, p. 109004, 2020.

\bibitem{b41}
J.~Qi, J.~Guo, M.~Wang, C.~Wu, and Z.~Ma, ``Formation tracking and obstacle
  avoidance for multiple quadrotors with static and dynamic obstacles,''
  \emph{IEEE Robotics and Automation Letters}, vol.~7, no.~2, pp. 1713--1720,
  2022.

\bibitem{b7}
W.~Ding, W.~Gao, K.~Wang, and S.~Shen, ``An efficient b-spline-based
  kinodynamic replanning framework for quadrotors,'' \emph{IEEE Transactions on
  Robotics}, vol.~35, no.~6, pp. 1287--1306, 2019.

\bibitem{b42}
G.~Sartoretti, J.~Kerr, Y.~Shi, G.~Wagner, T.~S. Kumar, S.~Koenig, and
  H.~Choset, ``{PRIMAL}: Pathfinding via reinforcement and imitation
  multi-agent learning,'' \emph{IEEE Robotics and Automation Letters}, vol.~4,
  no.~3, pp. 2378--2385, 2019.

\bibitem{b43}
C.~E. Luis, M.~Vukosavljev, and A.~P. Schoellig, ``Online trajectory generation
  with distributed model predictive control for multi-robot motion planning,''
  \emph{IEEE Robotics and Automation Letters}, vol.~5, no.~2, pp. 604--611,
  2020.

\bibitem{b18}
G.~V{\'a}s{\'a}rhelyi, C.~Vir{\'a}gh, G.~Somorjai, T.~Nepusz, A.~E. Eiben, and
  T.~Vicsek, ``Optimized flocking of autonomous drones in confined
  environments,'' \emph{Science Robotics}, vol.~3, no.~20, p. eaat3536, 2018.

\bibitem{b44}
R.~T. Rodrigues, M.~Basiri, A.~P. Aguiar, and P.~Miraldo, ``Low-level active
  visual navigation: Increasing robustness of vision-based localization using
  potential fields,'' \emph{IEEE Robotics and Automation Letters}, vol.~3,
  no.~3, pp. 2079--2086, 2018.

\bibitem{b8}
L.~Wang, A.~D. Ames, and M.~Egerstedt, ``Safety barrier certificates for
  collisions-free multirobot systems,'' \emph{IEEE Transactions on Robotics},
  vol.~33, no.~3, pp. 661--674, 2017.

\bibitem{b17}
Q.~Quan, Y.~Gao, and C.~Bai, ``Distributed control for a robotic swarm to pass
  through a curve virtual tube,'' \emph{Robotics and Autonomous Systems}, p.
  104368, 2023.

\bibitem{b21}
Q.~Quan, R.~Fu, M.~Li, D.~Wei, Y.~Gao, and K.-Y. Cai, ``Practical distributed
  control for {VTOL} {UAV}s to pass a virtual tube,'' \emph{IEEE Transactions
  on Intelligent Vehicles}, vol.~7, no.~2, pp. 342--353, 2021.

\bibitem{b31}
Y.~Wan, J.~Tang, and S.~Lao, ``Distributed conflict-detection and resolution
  algorithm for {UAV} swarms based on consensus algorithm and strategy
  coordination,'' \emph{IEEE Access}, vol.~7, pp. 100\,552--100\,566, 2019.

\bibitem{b32}
C.~Yan, C.~Wang, X.~Xiang, K.~H. Low, X.~Wang, X.~Xu, and L.~Shen,
  ``Collision-avoiding flocking with multiple fixed-wing {UAV}s in
  obstacle-cluttered environments: A task-specific curriculum-based madrl
  approach,'' \emph{IEEE Transactions on Neural Networks and Learning Systems},
  2023.

\bibitem{b33}
C.~Sun, J.~Leng, and F.~Sun, ``A fast optimal speed planning system in arterial
  roads for intelligent and connected vehicles,'' \emph{IEEE Internet of Things
  Journal}, vol.~9, no.~20, pp. 20\,295--20\,307, 2022.

\bibitem{b34}
J.~Villagra, V.~Milan{\'e}s, J.~P{\'e}rez, and J.~Godoy, ``Smooth path and
  speed planning for an automated public transport vehicle,'' \emph{Robotics
  and Autonomous Systems}, vol.~60, no.~2, pp. 252--265, 2012.

\bibitem{b35}
L.~L{\"u}tzow, Y.~Meng, A.~C. Armijos, and C.~Fan, ``Density planner:
  Minimizing collision risk in motion planning with dynamic obstacles using
  density-based reachability,'' in \emph{2023 IEEE International Conference on
  Robotics and Automation (ICRA)}.\hskip 1em plus 0.5em minus 0.4em\relax IEEE,
  2023, pp. 7886--7893.

\bibitem{b36}
C.~Sinigaglia, A.~Manzoni, F.~Braghin, and S.~Berman, ``Robust optimal density
  control of robotic swarms,'' \emph{arXiv preprint arXiv:2205.12592}, 2022.

\bibitem{b12}
P.~Mao and Q.~Quan, ``Making robotics swarm flow more smoothly: A regular
  virtual tube model,'' in \emph{2022 IEEE/RSJ International Conference on
  Intelligent Robots and Systems (IROS)}.\hskip 1em plus 0.5em minus
  0.4em\relax IEEE, 2022, pp. 4498--4504.

\bibitem{b26}
A.~M. Rezende, V.~M. Gon{\c{c}}alves, G.~V. Raffo, and L.~C. Pimenta, ``Robust
  fixed-wing {UAV} guidance with circulating artificial vector fields,'' in
  \emph{2018 IEEE/RSJ International Conference on Intelligent Robots and
  Systems (IROS)}.\hskip 1em plus 0.5em minus 0.4em\relax IEEE, 2018, pp.
  5892--5899.

\bibitem{b45}
R.~Olfati-Saber, ``Near-identity diffeomorphisms and exponential
  $\epsilon$-tracking and $\epsilon$-stabilization of first-order nonholonomic
  \emph{SE(2)} vehicles,'' in \emph{Proceedings of the 2002 American Control
  Conference (IEEE Cat. No. ch37301)}, vol.~6.\hskip 1em plus 0.5em minus
  0.4em\relax IEEE, 2002, pp. 4690--4695.

\bibitem{b24}
E.~Soria, F.~Schiano, and D.~Floreano, ``Predictive control of aerial swarms in
  cluttered environments,'' \emph{Nature Machine Intelligence}, vol.~3, no.~6,
  pp. 545--554, 2021.

\bibitem{b23}
S.~Wilson, P.~Glotfelter, L.~Wang, S.~Mayya, G.~Notomista, M.~Mote, and
  M.~Egerstedt, ``The robotarium: Globally impactful opportunities, challenges,
  and lessons learned in remote-access, distributed control of multirobot
  systems,'' \emph{IEEE Control Systems Magazine}, vol.~40, no.~1, pp. 26--44,
  2020.

\end{thebibliography}

\end{document}